\pdfoutput=1

\documentclass[11pt]{article}

\usepackage{acl}

\usepackage{times}
\usepackage{latexsym}

\usepackage[T1]{fontenc}

\usepackage[utf8]{inputenc}

\usepackage{microtype}

\title{Revisiting Generative Commonsense Reasoning: A Pre-Ordering Approach}

\author{Chao Zhao$^{1}$ \qquad Faeze Brahman$^{2}$ \qquad Tenghao Huang$^{1}$ \qquad Snigdha Chaturvedi$^{1}$ \\
\texttt{\{zhaochao, tenghao,  snigdha\}@cs.unc.edu}\qquad \texttt{faezeb@allenai.org}\\$^{1}$ UNC Chapel Hill \qquad $^{2}$ Allen Institute for AI}

\usepackage{amsmath,bm}
\usepackage{amssymb}
\usepackage{mathtools}
\usepackage[flushleft]{threeparttable}
\usepackage{soul}
\usepackage[normalem]{ulem} %
\usepackage{url}
\usepackage{paralist}
\usepackage{rotating,makecell}
\usepackage{tabularx}
\usepackage{array}
\usepackage{lipsum}
\usepackage{caption}
\usepackage{subcaption}

\usepackage{tabu}
\usepackage{hhline}
\usepackage{tabularx}
\usepackage{multirow,booktabs}
\newcommand{\RNum}[1]{\uppercase\expandafter{\romannumeral #1\relax}}
\usepackage{arydshln}
\usepackage[export]{adjustbox}

\usepackage{color}
\usepackage{enumitem}%
\usepackage[colorinlistoftodos,prependcaption,textsize=tiny]{todonotes}
\usepackage{xargs}                      %
\usepackage{xcolor}  %

\definecolor{red}{RGB}{189, 26, 26}
\definecolor{green}{RGB}{228, 249, 189}
\definecolor{Green}{RGB}{141, 213, 134}
\definecolor{purple}{RGB}{235, 219, 246}
\definecolor{pink}{RGB}{250, 221, 236}
\definecolor{smt}{RGB}{195, 62, 0}
\definecolor{nmt}{RGB}{252, 133, 0}
\definecolor{gmt}{RGB}{0, 98, 51}
\definecolor{pipe}{RGB}{0,128,255}

\definecolor{Orange}{RGB}{230,130,0}
\usepackage[normalem]{ulem}
\newcommand{\GG}[1]{}
\usepackage{float}

\newcommandx{\improvement}[2][1=]{\todo[linecolor=red,backgroundcolor=red!25,bordercolor=red,#1]{#2}}

\begin{document}
\maketitle
\begin{abstract}
Pre-trained models (PTMs) have lead to great improvements in natural language generation (NLG). However, it is still unclear how much commonsense knowledge they possess. With the goal of evaluating commonsense knowledge of NLG models, recent work has proposed the problem of generative commonsense reasoning, e.g., to compose a logical sentence given a set of unordered concepts. Existing approaches to this problem hypothesize that PTMs lack sufficient parametric knowledge for this task, which can be overcome by introducing external knowledge or task-specific pre-training objectives. Different from this trend, we argue that PTM's \textit{inherent} ability for generative commonsense reasoning is underestimated due to the order-agnostic property of its input. In particular, we hypothesize that the order of the input concepts can affect the PTM's ability to utilize its commonsense knowledge. To this end, we propose a pre-ordering approach to elaborately manipulate the order of the given concepts before generation. Experiments show that our approach can outperform the more sophisticated models that have access to a lot of external data and resources. 
\end{abstract}

\section{Introduction}
\begin{figure}[!t]
    \centering
    \includegraphics[width=1\linewidth]{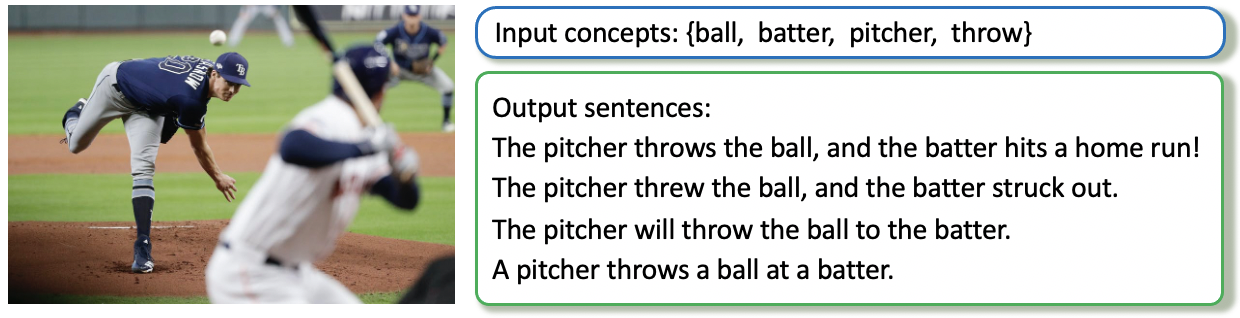}
    \caption{Example of the GCR task. }
    \label{fig:commongen}
\end{figure}

Pre-trained models (PTMs), such as BART \cite{lewis2020bart} and T5 \cite{raffel2020exploring}, have achieved significant progress in many natural language generation tasks. However, their ability to reason with common sense while generating text is questionable. To push research in this direction, \citet{lin2020commongen} proposed the task of generative commonsense reasoning (GCR), where the goal is to compose a fluent and rational sentence from a set of concepts.
Figure \ref{fig:commongen} shows an example of this problem. 
To achieve this goal, the model must do commonsense reasoning to build connections between the given concepts and produce a logically sound sentence (e.g., it is the \textit{pitcher} who throws the ball to the \textit{batter} rather than the other way). %

Prior works hypothesize that the vanilla PTMs are not capable of solving this challenging task \cite{liu2020kg, fan2020enhanced, zhou2021pretraining} partly because their self-supervised objectives do not explicitly capture the relational commonsense knowledge~\cite{zhou2021pretraining}.
These works enhance the PTMs' performance by explicitly introducing knowledge during fine-tuning or implicitly teaching the model during further pre-training. 
However, we observe that in some cases, even without external knowledge, PTMs can create reasonable output for this task, indicating that PTMs may already have the commonsense reasoning ability to some degree. Therefore the challenge turns out to be how to make it easier for PTMs to fully utilize the inherent commonsense knowledge.

One potential solution of this challenge is to make the order of input concepts more natural and aligned with commonsense. For example, in Figure \ref{fig:commongen}, taking \{\textit{pitcher}, \textit{throw}, \textit{ball}, \textit{batter}\} as the input is better than \{\textit{batter}, \textit{throw}, \textit{ball}, \textit{pitcher}\}, since the order of concepts in the former input is more close to that in the outputs. Models that are not pre-trained, such as LSTM and GRU, prefer a pre-ordering of input tokens to align them with the (expected) output \cite{VinyalsBK15, 10.1162/COLI_a_00245}. For PTMs, recent works \cite{kale2020text, ribeiro2020investigating,hoyle2020promoting} show that they can achieve reasonable performance on graph-to-text tasks without pre-ordering. However, the impact of pre-ordering on PTMs, in general, is not well analyzed.

In this work, we revisit PTMs' ability of generative commonsense reasoning \textit{without access to external knowledge or task-specific pre-training}. We choose BART and T5, two state-of-the-art PTMs, as our underlying models.
To analyze the utility of pre-ordering the concepts on models' performance, we introduce \textbf{Planned-BART} and \textbf{Planned-T5} to manipulate the input concept order before generation, which helps to make the order of input concepts more natural (more close to the order of concepts in the output sentence).
We experimentally show that via pre-ordering, Planned-BART and Planned-T5 exceed the more sophisticated models that have access to external knowledge or training data. It
indicates that PTM's inherent ability for generative commonsense reasoning was underestimated while a simple pre-ordering step can help PTMs better use this ability.

\section{Related Works}
\subsection{Generative Commonsense Reasoning}
\label{subsec::gcr}
There are two major approaches to enhance the vanilla PTM's ability of commonsense reasoning on generation. The first approach is to introduce explicit knowledge from external sources such as ConceptNet \cite{liu2020kg} and retrieved prototypes \cite{fan2020enhanced,wang-etal-2021-retrieval-enhanced}, which can facilitate GSR by either building connections between related concepts or providing adjunct words for the input.
The second approach is to explicitly teach models to reason over the concepts via new pre-training objectives \cite{zhou2021pretraining}. Different from these works, we examine PTMs’ inherent ability of GSR without the help of external knowledge or task-specific pre-training.

\subsection{Sequence Pre-Ordering}
Previous works have shown that the pre-ordering of input sequence can improve the task of graph-to-text generation \cite{moryossef2019step,zhao2020bridging}, but they use non-pre-trained LSTM and the pre-ordering methods rely on rich structural information from the input. We instead focus on PTMs and non-structural input. For PTMs, \citet{hessel2021effective} and \citet{sinha2021masked} show that PTMs are resilient to shuffling the order of input tokens on the tasks of natural language understanding, but they didn't study the generation problem. \citet{hoyle2020promoting} show that a suitable pre-ordering can improve the generation quality. However, they didn’t provide a general pre-ordering method for the problem of keywords-to-text generation.

\section{Generative Commonsense Reasoning}

\subsection{Task Formalization}
Given a set of lemmatized tokens representing  concepts $\mathcal{X} = \{x_1, \cdots, x_m\}$, where each $x_i$ can be a noun or a verb, the goal is to generate a fluent and grammatically correct English sentence $\mathbf{y}=\{y_1, \cdots, y_n\}$ such that it contains all of the concepts in $\mathcal{X}$.
The task does not require $x_i$ to have the same morphological form as it appears in $\mathbf{y}$. Figure \ref{fig:commongen} shows an example of the task. Note that $\mathcal{X}$ is an unordered set of concepts. We refer to a permutation of $\mathcal{X}$ as a \textit{Plan} of the concept set. For a given output sentence $\mathbf{y}$, we re-order $\mathcal{X}$ to make the concepts have the same order as those in $\mathbf{y}$ and call it as the \textit{Skeleton} of $\mathbf{y}$.
Note that skeletons are associated with the outputs while plans are determined before generation.
We refer to the plans which are identical to the references' skeletons as \textit{Oracle Plans}.

\begin{table*}
\small
	\centering
	\begin{tabular}{l|c|c|c|c|c|c} 
\hline 
Model $\backslash$ Metrics & ROUGE-2/L &  BLEU-3/4  & METEOR & CIDEr & SPICE & Coverage \\
\hline 
BART \cite{lin2020commongen} & ${22.23}$\quad$41.98$ & $36.3$\quad$26.3$ & ${30.9}$ & $13.92$ & ${30.6}$ & ${97.35}$ \\
EKI-BART \cite{fan2020enhanced} & ${24.36}$\quad${45.42}$ & ${42.9}$\quad${32.1}$  & ${32.0}$ & ${16.80}$ & ${32.5}$ & - \\
KG-BART \cite{liu2020kg} & ${23.38}$\quad${44.54}$ & ${42.1}$\quad${30.9}$ & ${32.4}$ & ${16.83}$ & ${32.7}$ & ${98.68}$\\
Planned-BART (Ours) & ${24.97}$\quad${46.13}$ & ${44.8}$\quad${34.1}$ & ${32.9}$ & ${17.47}$ & ${33.1}$ & ${98.99}$\\
\hline
T5 \cite{lin2020commongen} & ${22.01}$ \quad ${42.97}$ & ${39.0}$ \quad ${28.6}$ & ${30.1}$ & ${14.96}$ & ${31.6}$ & ${95.29}$ \\
CALM \cite{zhou2021pretraining} & - \quad\quad\quad - & \quad - \quad $\mkern9mu {29.5}$ & ${31.9}$ & ${15.61}$ & ${33.2}$ & - \\
RE-T5 \cite{wang-etal-2021-retrieval-enhanced} & - \quad\quad\quad - & - \quad\quad - & - & - & ${34.3}$ & - \\
Planned-T5 (Ours) & ${24.07}$\quad${46.11}$ & ${44.6}$\quad${33.7}$ & ${32.8}$ & ${17.60}$ & ${34.0}$ & ${98.60}$\\

 \hline
\hline Human Performance & $48.88$ \quad $63.79$ & $48.2$ \quad $44.9$ & $36.2$ & $43.53$ & $63.5$ & $99.31$ \\
\hline
\end{tabular}

	\caption{\label{tab:res_gen} Automatic evaluation of generation quality. We compare our methods with pre-train- or knowledge-enhanced baselines. Our best model outperforms previous models on all automatic measures. The only exception is RE-T5, which uses both external knowledge and pre-training (with 7 times larger training data). }
\end{table*}

We use BART and T5, two state-of-the-art PTMs, as the underlying generation models. Both models are based on the Transformer architecture~\cite{vaswani2017attention}.  Similar to other sequence-to-sequence models, they receive $\mathbf{x}=\{x_1, \cdots, x_m\}$ as input, and model the probability of the output sequence $\mathbf{y}=\{y_1, \cdots, y_n\}$ as: 
\begin{equation}
p(\mathbf{y} \mid \mathbf{x} ; \boldsymbol{\theta})=\prod_{t=1}^{|\mathbf{y}|} p\left(y_{t} \mid y_{1: t-1}, \mathbf{x} ; \boldsymbol{\theta}\right).
\label{eq:prob_y}
\end{equation}

\subsection{Planned Model}
\label{subsec::planned-bart}
To fine-tune PTMs on this task, previous works regard the input as an unordered set and use its random linearization as the input in both training and inference phases. Although it is trained in an order-agnostic setting, PTMs are naturally position-sensitive models because the same input words in different permutations have different positional representations.

Leveraging this property, we introduce Planned-BART and Planned-T5 to make both models aware of the input order by regarding the input as an ordered sequence.
To order the input concepts properly, in the training phase, we re-order the concepts according to the corresponding oracle plan. That is, we force the order of concepts in both input and output sequences to be identical during training, which can better help the model utilize its inherent commonsense reasoning capabilities. 
In the inference phase, the oracle plans of concepts are unavailable. We instead obtain the plan using a  \textit{Planner}. Leveraging the power of PTMs, the planner is a vanilla BART or T5 model, which is fine-tuned on unordered (randomly linearized) input and produces a sentence as output. The skeleton of the planner's output forms the plan for planned models.

\section{Experiments}
\label{sec::exp}

\subsection{Dataset and Evaluation}
We conduct experiments \footnote{Code is available at \url{https://github.com/zhaochaocs/Planned-PTM}} on the \textsc{CommonGen} dataset \cite{lin2020commongen}, which contains 35k concepts-sentence pairs for training/validation/test.
To build concepts-reference pairs, \textsc{CommonGen} first collects frequently co-occurring concepts from image captions. Each concept-set contains three to five concepts. The references in the training set are original captions while those in the validation and test sets are collected by crowd-sourcing. 

The quality of the generated text is evaluated through several automatic metrics such as BLEU \cite{papineni2002bleu}, ROUGE \cite{lin2004rouge}, METEOR \cite{banerjee2005meteor}, CIDEr \cite{vedantam2015cider}, and SPICE \cite{anderson2016spice}. We also report Coverage \cite{lin2020commongen}, which is the average percentage of input concepts that are present in the output sentences.

\subsection{Results}
We compare the performance of our pre-ordered method with the unordered BART and T5, as well as two knowledge-enhanced BART models: EKI-BART and KG-BART, and two T5 models enhanced by further pre-training: CALM and RE-T5. 
Table \ref{tab:res_gen} lists the results of automatic measures. The training details can be found in Appendix \ref{app::train}.

Our Planned-BART and Planned-T5 models outperform vanilla BART and T5 models, demonstrating that pre-ordering the input helps PTMs in effectively leveraging their inherent commonsense knowledge. Our models also outperform three out of four baselines that use external knowledge or pre-training objectives. The only exception is RE-T5, which is further pre-trained.%
This indicates that PTMs inherently contain a lot of commonsense knowledge that needs to be first utilized before bringing in information from external sources.
 
To further explore the potential of the pre-ordering method, we conduct another experiment to investigate the impact of concept orders on generation quality. Given a test concept set, we feed all of its permutations to either BART or Planned-BART to generate sentences. We then rank the sentences according to their probabilities in Equation \noindent \ref{eq:prob_y} and pick the most probable sentence as the final output. We refer to the methods using this strategy as BART \textsubscript{Rank} and Planned-BART \textsubscript{Rank}, respectively. 
Note that the ranking method is computationally inefficient. In this work, we only use these models to provide an estimate of the upper bound on the performance of the pre-ordering method.

As shown in Table \ref{tab::rank_gen}, the performance of Planned-BART is close to its ranking variant. This demonstrates the effectiveness of our planning strategy -- it helps Planned-BART achieve a performance comparable to the upper bound at a much lower computational overhead. We also observe that Planned-BART \textsubscript{Rank} achieves better scores than BART \textsubscript{Rank}. This is because Planned-BART is trained on oracle plans, which helps it in better utilizing its inherent commonsense knowledge.

\begin{table}
\small
		\centering
		\setlength{\tabcolsep}{0.5em} %
		
\begin{tabular}{l|c|c|c|c|c} 
\hline 
Model $\backslash$ Metrics & R-2 & B-4 & M & C & S \\
\hline 
Planned-BART & 24.97 & 34.1 & 32.9 & 17.47 & 33.1 \\
BART \textsubscript{Rank} & 24.31 & 33.0 & 33.0 & 17.39 & 33.2 \\
Planned-BART \textsubscript{Rank} & 25.04 & 35.0 & 33.3 & 17.89 & 33.6 \\
\hline
\end{tabular}

		\caption{\label{tab::rank_gen} Evaluation of Planned-BART and ranking models on ROUGE-2, BLEU-4, METEOR, CIDEr, and SPICE. }
\end{table}

\subsection{Human Evaluation}
We randomly select 100 test instances and evaluate the generation quality of a system according to Rationality, Fluency, and Succinctness as in \citet{liu2020kg}. We conduct a pairwise comparison between Planned-BART \textsubscript{Rank} (the best model) with our three other methods.%
For each test instance, we obtain the output sentences from two different models, and then ask three workers on Amazon Mechanical Turk to compare the two sentences according to the three measures listed above. More details can be found in Appendix \ref{app::eval}.

Table \ref{tab::res_human_eval} lists the results, where negative scores indicate worse performance compared with Planned-BART \textsubscript{Rank}. 
The original BART performs the worst on all measures, while Planned-BART achieves closer quality to BART \textsubscript{Rank} and Planned-BART \textsubscript{Rank}. These results are consistent with those of automatic evaluations and support our claim that Planned-BART can be a reasonable trade-off between performance and efficiency.

\begin{table}
\small
		\centering
		\begin{tabular}{l|c|c|c} 
\hline 
Model $\backslash$ Metrics & \textsc{Ration} & \textsc{Fluency}  &  \textsc{Succinct} \\
\hline 
BART & -0.38 & -0.33 & -0.56 \\
BART \textsubscript{Rank}  & -0.10 & 0.04 & -0.13 \\
Planned-BART  & -0.29 & -0.19 & -0.17 \\
\hline
\end{tabular}

		\caption{\label{tab::res_human_eval} Results of human evaluation on rationality, fluency, and succinctness. We report the pair-wise scores between Planned-BART \textsubscript{Rank} (the best model) with three other models. Negative scores indicates worse performance compared with Planned-BART \textsubscript{Rank}.%
		}
\end{table}

\section{Analysis}
In this section, we analyze the impact of input permutation on the model and the generated sentences.

\subsection{Permutation Invariance}
\label{subsec::invariance}
We first examine how the output changes when the original BART (which we refer to as Unordered-BART for clarity) and Planned-BART receive all possible permutations of concepts as input. We compare the skeleton of Planned-BART's outputs with the input plans and find that the output skeleton is consistent with the order of input concept in 94\% of the cases, which is as expected.

In contrast, for Unordered-BART, we find that for 61\% of the permutations, it can organize the concepts in one particular order irrespective of the input order. More details are provided in Appendix \ref{app::invariance}.
This observation suggests that Unordered-BART is permutation-invariant to the input to some degree. However, it is difficult for the model to be entirely insensitive to the input permutation, which explains the performance difference between BART and BART \textsubscript{Rank}: a ranking strategy helps select a more suitable permutation of input and can therefore improve the generation quality. 

\subsection{Impact of Permutation on Encoding}
The observations in Section \ref{subsec::invariance} prompt a question about how Unordered-BART and Planned-BART have different behaviors when receiving input permutation. Here, we explore this question by studying the impact of input permutation on the model encoder, especially the global attention distributions and the local attention strength between certain word pairs.

One possible reason for the permutation invariance of Unordered-BART is that although different input plans have different positional embeddings and may affect hidden states of the lower layers, the encoder can build stable association among tokens at the higher layers, alleviating the disturbance from positional embeddings. For example, the model may know that people should ``ride a bike on trail'' even when the concept order is \{ride, trail, bike\}. We measure the word association inside the encoder using the strength of attention weights between concepts.

\begin{figure}[t]
     \centering
     \begin{subfigure}[b]{0.23\textwidth}
         \includegraphics[width=0.95\textwidth, left]{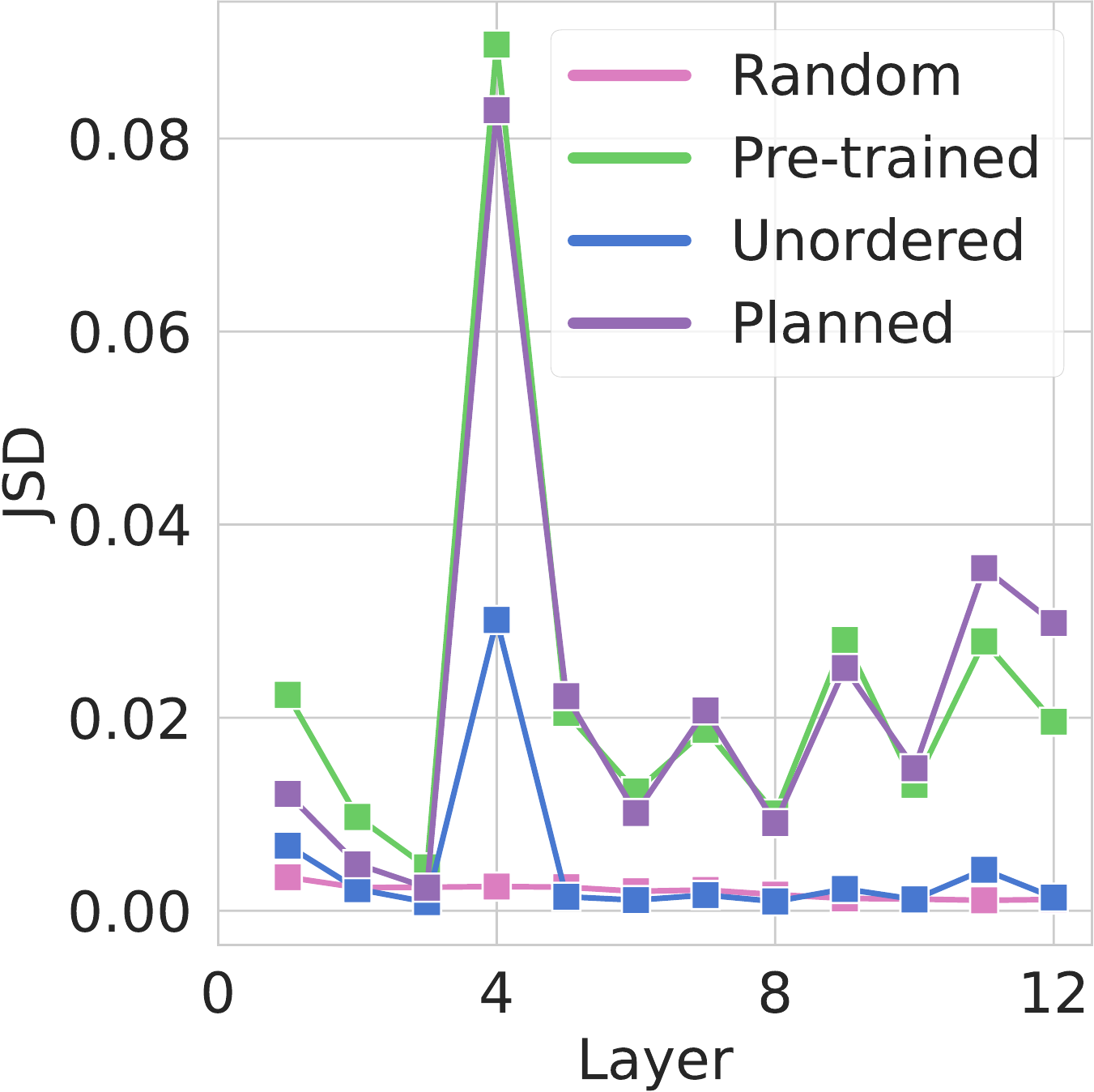}
     \end{subfigure}
     \begin{subfigure}[b]{0.23\textwidth}
         \centering
\includegraphics[width=0.95\textwidth, right]{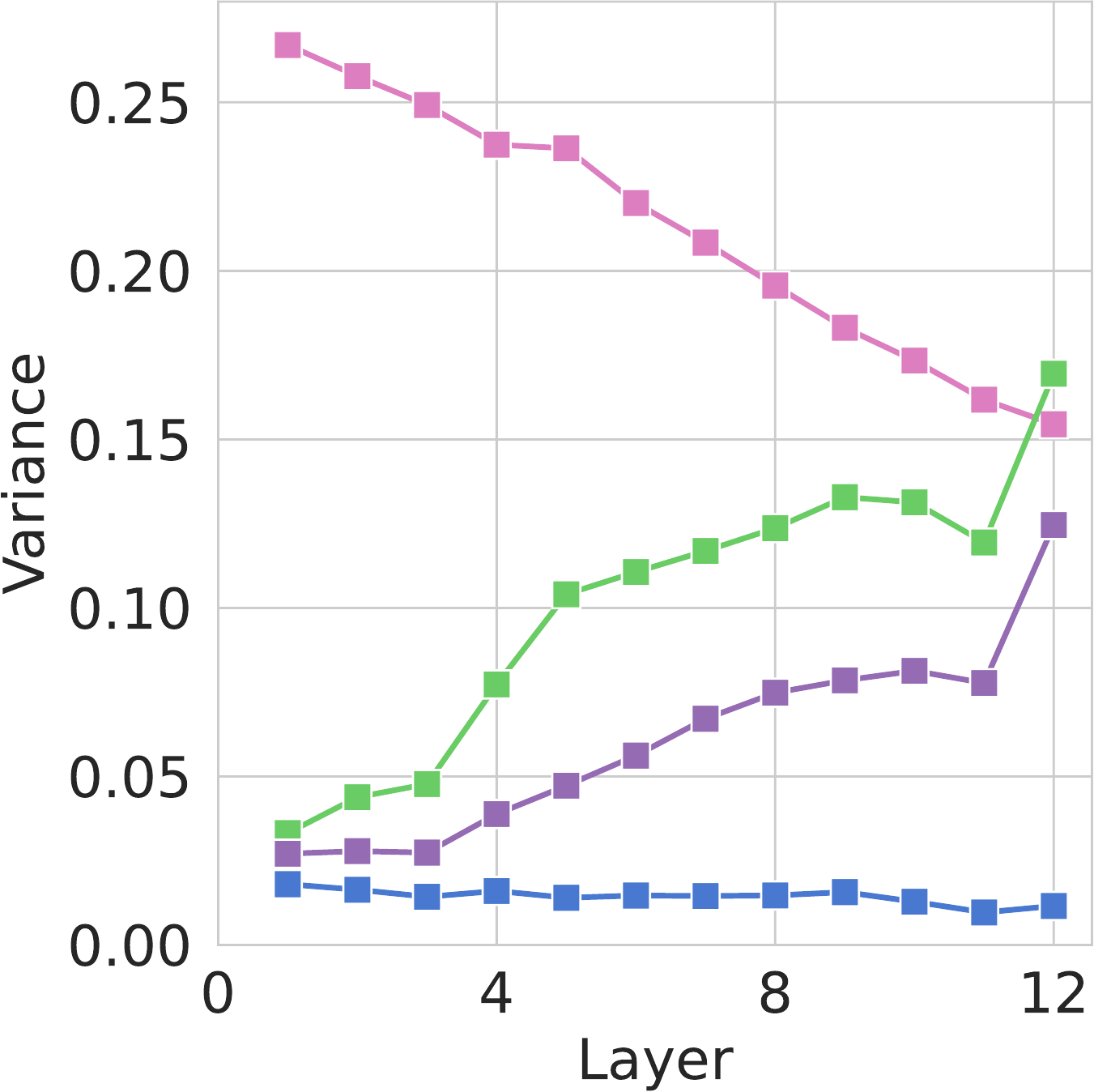}
     \end{subfigure}
    \caption{Left: JS divergence of encoder attention distributions w.r.t. the input permutations. Right: the variance of encoder hidden states w.r.t. input permutations. }
\label{fig::var}
\end{figure}

To verify our assumption, we calculate (i) %
the Jensen–Shannon divergence (JSD) of the encoder attention distributions w.r.t. all possible permutations of the input, and (ii) the variance of encoder hidden states w.r.t. the input permutations.
Figure \ref{fig::var} shows the layer-wise JSD and variance averaged over the test set. For comparison, we also include the results from a randomly-initialized BART and the pre-trained BART (without fine-tuning).

From Figure \ref{fig::var} we observe that the attention distributions of Planned-BART have high JS divergence at each layer, and have a similar trend compared with that of Pre-trained BART. It indicates that attention distributions of these two models are affected by the input permutation, which is expected since their input is well ordered during pre-training or fine-tuning. %
As a result, the variances of hidden states on both models increase with a growth in layer depth. In contrast, the JS divergence of attention in Unordered-BART gets close to 0 starting from layer 2 and becomes similar to that of the randomly initialized model. It indicates that the encoder can assign stable attention distributions to the input despite the difference in permutation. Because of this, the variances of hidden states on these two models decrease as the layer goes deeper. It partially explains the permutation invariance of Unordered-BART. %
We also noticed a substantial negative correlation between the variance of the encoder output and the percentage of the mode sentence (Spearman’s $\rho = - 0.435$), which supports our explanation.

In addition to the analysis of global attention distribution, we also investigate local attention patterns, i.e., whether the attention weights between concept tokens can reflect their commonsense relations. More details are listed in Appendix \ref{app::local_attn}. We find that compared with a randomly-initialized BART, the pre-trained BART is better at tracking commonsense relations of concepts despite input permutation, and fine-tuning can further strengthen this capability. We also find that the model heavily relies on 
the tracking ability when generating texts. It demonstrates that BART has the commonsense reasoning ability to some degree, and it is reasonable to leverage the output of Unordered-BART to obtain the plan for planned-BART.

\subsection{Impact of Permutation on Decoding}
In this section, we discuss how the input permutation can affect the quality of decoding output. Particularly, we show that reasonable planning can create less repetitive and more diverse output.

First, we find that the unordered models suffer from the repetition of content in the output. For example, in 34.2\% of test cases, there is at least one concept that appears more than once in the output of the unordered BART. However, this percentage decreased to 3.2\% for the output of Planned-BART. 
It is because the decoder of Planned-BART can assign attention weights monotonically to the input, and reduce the repetition caused by re-attending the previous concepts. In Appendix \ref{app::rep}, we provide a visualization of how the input order can impact the attention weights during decoding. 

\begin{figure}
    \centering
    \includegraphics[width=0.6\linewidth]{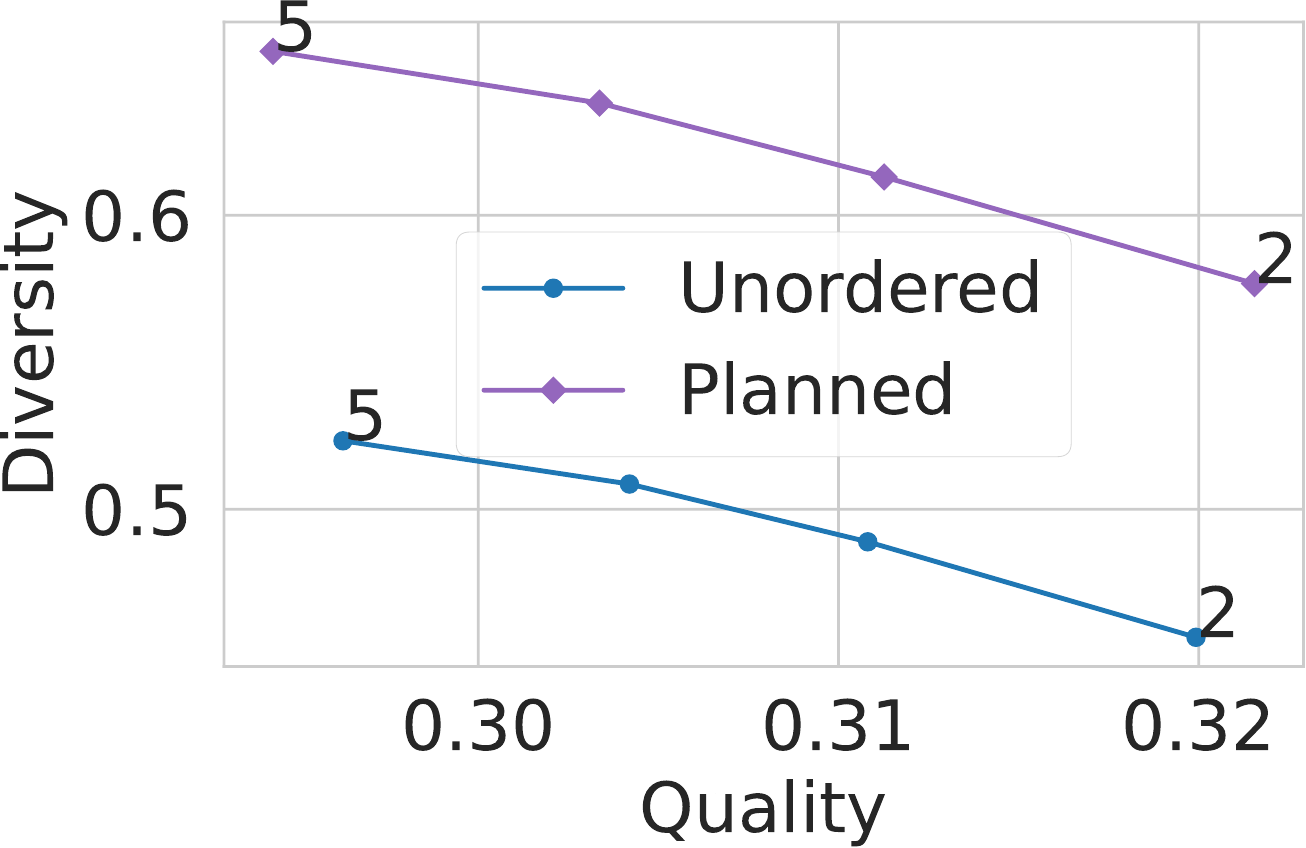}
    \caption{\label{fig::div} The quality-diversity plot of Unordered-BART \textsubscript{Rank} and Planned-BART \textsubscript{Rank}. }
\end{figure}

Second, the order consistency between inputs and outputs in Planned-BART also allows us to have more control over output skeletons by adjusting the input concept order. 
Different orders can help the encoder to capture diverse commonsense relations between concepts and create diverse outputs. While unnatural diversity may hurt generation quality, we use SPICE as the measure for quality and BLEU-based discrepancy \cite{shu2019generating} for diversity, and evaluate the performance of Unordered-BART \textsubscript{Rank} and Planned-BART \textsubscript{Rank} by selecting the top 2 to 5 candidates as outputs. Figure \ref{fig::div} shows the quality-diversity plot of two models. It indicates that with little degradation of generation quality, Planned-BART can create more diverse output than Unordered-BART. We show an example in Appendix \ref{app::div}.

\section{Conclusion}
In this work, we revisit the PTM's inherent ability of generative commonsense reasoning. We use BART and T5 as underlying generators and propose their planned variants to manipulate the order of the given concepts before generation. Experiments on \textsc{CommonGen} dataset demonstrate that this simple pre-ordering approach can outperform the previous pre-trained or knowledge-enhanced models. 
Besides that, planned models can leverage the pre-ordered concepts to create more succinct and diverse sentences. In conclusion, our work suggests that PTM's inherent ability for generative commonsense reasoning is underestimated due to the unordered input, and the pre-ordering step can help PTMs to improve the generation quality.

\bibliography{custom}

\begin{thebibliography}{29}
\expandafter\ifx\csname natexlab\endcsname\relax\def\natexlab#1{#1}\fi

\bibitem[{Anderson et~al.(2016)Anderson, Fernando, Johnson, and
  Gould}]{anderson2016spice}
Peter Anderson, Basura Fernando, Mark Johnson, and Stephen Gould. 2016.
\newblock Spice: Semantic propositional image caption evaluation.
\newblock In \emph{European conference on computer vision}, pages 382--398.
  Springer.

\bibitem[{Banerjee and Lavie(2005)}]{banerjee2005meteor}
Satanjeev Banerjee and Alon Lavie. 2005.
\newblock \href {https://aclanthology.org/W05-0909} {{METEOR}: An automatic
  metric for {MT} evaluation with improved correlation with human judgments}.
\newblock In \emph{Proceedings of the {ACL} Workshop on Intrinsic and Extrinsic
  Evaluation Measures for Machine Translation and/or Summarization}, pages
  65--72, Ann Arbor, Michigan. Association for Computational Linguistics.

\bibitem[{Bisazza and Federico(2016)}]{10.1162/COLI_a_00245}
Arianna Bisazza and Marcello Federico. 2016.
\newblock \href {https://doi.org/10.1162/COLI_a_00245} {{A Survey of Word
  Reordering in Statistical Machine Translation: Computational Models and
  Language Phenomena}}.
\newblock \emph{Computational Linguistics}, 42(2):163--205.

\bibitem[{Clark et~al.(2019)Clark, Khandelwal, Levy, and
  Manning}]{clark2019does}
Kevin Clark, Urvashi Khandelwal, Omer Levy, and Christopher~D. Manning. 2019.
\newblock \href {https://doi.org/10.18653/v1/W19-4828} {What does {BERT} look
  at? an analysis of {BERT}{'}s attention}.
\newblock In \emph{Proceedings of the 2019 ACL Workshop BlackboxNLP: Analyzing
  and Interpreting Neural Networks for NLP}, pages 276--286, Florence, Italy.
  Association for Computational Linguistics.

\bibitem[{Fan et~al.(2020)Fan, Gong, Wei, Wang, Huang, Jiao, Huang, Duan, and
  Zhang}]{fan2020enhanced}
Zhihao Fan, Yeyun Gong, Zhongyu Wei, Siyuan Wang, Yameng Huang, Jian Jiao,
  Xuanjing Huang, Nan Duan, and Ruofei Zhang. 2020.
\newblock \href {https://doi.org/10.18653/v1/2020.coling-main.182} {An enhanced
  knowledge injection model for commonsense generation}.
\newblock In \emph{Proceedings of the 28th International Conference on
  Computational Linguistics}, pages 2014--2025, Barcelona, Spain (Online).
  International Committee on Computational Linguistics.

\bibitem[{Hessel and Schofield(2021)}]{hessel2021effective}
Jack Hessel and Alexandra Schofield. 2021.
\newblock \href {https://doi.org/10.18653/v1/2021.acl-short.27} {How effective
  is {BERT} without word ordering? implications for language understanding and
  data privacy}.
\newblock In \emph{Proceedings of the 59th Annual Meeting of the Association
  for Computational Linguistics and the 11th International Joint Conference on
  Natural Language Processing (Volume 2: Short Papers)}, pages 204--211,
  Online. Association for Computational Linguistics.

\bibitem[{Hoyle et~al.(2021)Hoyle, Marasovi{\'c}, and
  Smith}]{hoyle2020promoting}
Alexander~Miserlis Hoyle, Ana Marasovi{\'c}, and Noah~A. Smith. 2021.
\newblock \href {https://doi.org/10.18653/v1/2021.findings-acl.82} {Promoting
  graph awareness in linearized graph-to-text generation}.
\newblock In \emph{Findings of the Association for Computational Linguistics:
  ACL-IJCNLP 2021}, pages 944--956, Online. Association for Computational
  Linguistics.

\bibitem[{Htut et~al.(2019)Htut, Phang, Bordia, and Bowman}]{htut2019attention}
Phu~Mon Htut, Jason Phang, Shikha Bordia, and Samuel~R Bowman. 2019.
\newblock \href {https://arxiv.org/abs/1911.12246} {Do attention heads in bert
  track syntactic dependencies?}
\newblock \emph{arXiv preprint arXiv:1911.12246}.

\bibitem[{Kale and Rastogi(2020)}]{kale2020text}
Mihir Kale and Abhinav Rastogi. 2020.
\newblock \href {https://aclanthology.org/2020.inlg-1.14} {Text-to-text
  pre-training for data-to-text tasks}.
\newblock In \emph{Proceedings of the 13th International Conference on Natural
  Language Generation}, pages 97--102, Dublin, Ireland. Association for
  Computational Linguistics.

\bibitem[{Kingma and Ba(2015)}]{kingma2014adam}
Diederik~P. Kingma and Jimmy Ba. 2015.
\newblock \href {http://arxiv.org/abs/1412.6980} {Adam: {A} method for
  stochastic optimization}.
\newblock In \emph{3rd International Conference on Learning Representations,
  {ICLR} 2015, San Diego, CA, USA, May 7-9, 2015, Conference Track
  Proceedings}.

\bibitem[{Lewis et~al.(2020)Lewis, Liu, Goyal, Ghazvininejad, Mohamed, Levy,
  Stoyanov, and Zettlemoyer}]{lewis2020bart}
Mike Lewis, Yinhan Liu, Naman Goyal, Marjan Ghazvininejad, Abdelrahman Mohamed,
  Omer Levy, Veselin Stoyanov, and Luke Zettlemoyer. 2020.
\newblock \href {https://doi.org/10.18653/v1/2020.acl-main.703} {{BART}:
  Denoising sequence-to-sequence pre-training for natural language generation,
  translation, and comprehension}.
\newblock In \emph{Proceedings of the 58th Annual Meeting of the Association
  for Computational Linguistics}, pages 7871--7880, Online. Association for
  Computational Linguistics.

\bibitem[{Lin et~al.(2020)Lin, Zhou, Shen, Zhou, Bhagavatula, Choi, and
  Ren}]{lin2020commongen}
Bill~Yuchen Lin, Wangchunshu Zhou, Ming Shen, Pei Zhou, Chandra Bhagavatula,
  Yejin Choi, and Xiang Ren. 2020.
\newblock \href {https://doi.org/10.18653/v1/2020.findings-emnlp.165}
  {{C}ommon{G}en: A constrained text generation challenge for generative
  commonsense reasoning}.
\newblock In \emph{Findings of the Association for Computational Linguistics:
  EMNLP 2020}, pages 1823--1840, Online. Association for Computational
  Linguistics.

\bibitem[{Lin(2004)}]{lin2004rouge}
Chin-Yew Lin. 2004.
\newblock \href {https://aclanthology.org/W04-1013} {{ROUGE}: A package for
  automatic evaluation of summaries}.
\newblock In \emph{Text Summarization Branches Out}, pages 74--81, Barcelona,
  Spain. Association for Computational Linguistics.

\bibitem[{Liu et~al.(2021)Liu, Wan, He, Peng, and Yu}]{liu2020kg}
Ye~Liu, Yao Wan, Lifang He, Hao Peng, and Philip~S Yu. 2021.
\newblock Kg-bart: Knowledge graph-augmented bart for generative commonsense
  reasoning.
\newblock In \emph{Proceedings of the AAAI Conference on Artificial
  Intelligence}, volume~35.

\bibitem[{Michel et~al.(2019)Michel, Levy, and Neubig}]{NEURIPS2019_2c601ad9}
Paul Michel, Omer Levy, and Graham Neubig. 2019.
\newblock \href
  {https://proceedings.neurips.cc/paper/2019/hash/2c601ad9d2ff9bc8b282670cdd54f69f-Abstract.html}
  {Are sixteen heads really better than one?}
\newblock In \emph{Advances in Neural Information Processing Systems 32: Annual
  Conference on Neural Information Processing Systems 2019, NeurIPS 2019,
  December 8-14, 2019, Vancouver, BC, Canada}, pages 14014--14024.

\bibitem[{Moryossef et~al.(2019)Moryossef, Goldberg, and
  Dagan}]{moryossef2019step}
Amit Moryossef, Yoav Goldberg, and Ido Dagan. 2019.
\newblock {\GG{1}}step-by-step: Separating planning from realization in neural
  data-to-text generation.
\newblock In \emph{Proceedings of the 2019 Conference of the North American
  Chapter of the Association for Computational Linguistics: Human Language
  Technologies, Volume 1 (Long and Short Papers)}, pages 2267--2277.

\bibitem[{Papineni et~al.(2002)Papineni, Roukos, Ward, and
  Zhu}]{papineni2002bleu}
Kishore Papineni, Salim Roukos, Todd Ward, and Wei-Jing Zhu. 2002.
\newblock \href {https://doi.org/10.3115/1073083.1073135} {{B}leu: a method for
  automatic evaluation of machine translation}.
\newblock In \emph{Proceedings of the 40th Annual Meeting of the Association
  for Computational Linguistics}, pages 311--318, Philadelphia, Pennsylvania,
  USA. Association for Computational Linguistics.

\bibitem[{Raffel et~al.(2020)Raffel, Shazeer, Roberts, Lee, Narang, Matena,
  Zhou, Li, and Liu}]{raffel2020exploring}
Colin Raffel, Noam Shazeer, Adam Roberts, Katherine Lee, Sharan Narang, Michael
  Matena, Yanqi Zhou, Wei Li, and Peter~J Liu. 2020.
\newblock Exploring the limits of transfer learning with a unified text-to-text
  transformer.
\newblock \emph{Journal of Machine Learning Research}, 21:1--67.

\bibitem[{Ribeiro et~al.(2021)Ribeiro, Schmitt, Sch{\"u}tze, and
  Gurevych}]{ribeiro2020investigating}
Leonardo F.~R. Ribeiro, Martin Schmitt, Hinrich Sch{\"u}tze, and Iryna
  Gurevych. 2021.
\newblock \href {https://doi.org/10.18653/v1/2021.nlp4convai-1.20}
  {Investigating pretrained language models for graph-to-text generation}.
\newblock In \emph{Proceedings of the 3rd Workshop on Natural Language
  Processing for Conversational AI}, pages 211--227, Online. Association for
  Computational Linguistics.

\bibitem[{Shu et~al.(2019)Shu, Nakayama, and Cho}]{shu2019generating}
Raphael Shu, Hideki Nakayama, and Kyunghyun Cho. 2019.
\newblock \href {https://doi.org/10.18653/v1/P19-1177} {Generating diverse
  translations with sentence codes}.
\newblock In \emph{Proceedings of the 57th Annual Meeting of the Association
  for Computational Linguistics}, pages 1823--1827, Florence, Italy.
  Association for Computational Linguistics.

\bibitem[{Sinha et~al.(2021)Sinha, Jia, Hupkes, Pineau, Williams, and
  Kiela}]{sinha2021masked}
Koustuv Sinha, Robin Jia, Dieuwke Hupkes, Joelle Pineau, Adina Williams, and
  Douwe Kiela. 2021.
\newblock \href {https://doi.org/10.18653/v1/2021.emnlp-main.230} {Masked
  language modeling and the distributional hypothesis: Order word matters
  pre-training for little}.
\newblock In \emph{Proceedings of the 2021 Conference on Empirical Methods in
  Natural Language Processing}, pages 2888--2913, Online and Punta Cana,
  Dominican Republic. Association for Computational Linguistics.

\bibitem[{Vaswani et~al.(2017)Vaswani, Shazeer, Parmar, Uszkoreit, Jones,
  Gomez, Kaiser, and Polosukhin}]{vaswani2017attention}
Ashish Vaswani, Noam Shazeer, Niki Parmar, Jakob Uszkoreit, Llion Jones,
  Aidan~N. Gomez, Lukasz Kaiser, and Illia Polosukhin. 2017.
\newblock \href
  {https://proceedings.neurips.cc/paper/2017/hash/3f5ee243547dee91fbd053c1c4a845aa-Abstract.html}
  {Attention is all you need}.
\newblock In \emph{Advances in Neural Information Processing Systems 30: Annual
  Conference on Neural Information Processing Systems 2017, December 4-9, 2017,
  Long Beach, CA, {USA}}, pages 5998--6008.

\bibitem[{Vedantam et~al.(2015)Vedantam, Zitnick, and
  Parikh}]{vedantam2015cider}
Ramakrishna Vedantam, C.~Lawrence Zitnick, and Devi Parikh. 2015.
\newblock \href {https://doi.org/10.1109/CVPR.2015.7299087} {Cider:
  Consensus-based image description evaluation}.
\newblock In \emph{{IEEE} Conference on Computer Vision and Pattern
  Recognition, {CVPR} 2015, Boston, MA, USA, June 7-12, 2015}, pages
  4566--4575. {IEEE} Computer Society.

\bibitem[{Vinyals et~al.(2016)Vinyals, Bengio, and Kudlur}]{VinyalsBK15}
Oriol Vinyals, Samy Bengio, and Manjunath Kudlur. 2016.
\newblock \href {http://arxiv.org/abs/1511.06391} {Order matters: Sequence to
  sequence for sets}.
\newblock In \emph{4th International Conference on Learning Representations,
  {ICLR} 2016, San Juan, Puerto Rico, May 2-4, 2016, Conference Track
  Proceedings}.

\bibitem[{Wang et~al.(2021)Wang, Liu, Zhu, Shou, Gong, Xu, and
  Zeng}]{wang-etal-2021-retrieval-enhanced}
Han Wang, Yang Liu, Chenguang Zhu, Linjun Shou, Ming Gong, Yichong Xu, and
  Michael Zeng. 2021.
\newblock \href {https://doi.org/10.18653/v1/2021.findings-acl.269} {Retrieval
  enhanced model for commonsense generation}.
\newblock In \emph{Findings of the Association for Computational Linguistics:
  ACL-IJCNLP 2021}, pages 3056--3062, Online. Association for Computational
  Linguistics.

\bibitem[{Wolf et~al.(2020)Wolf, Debut, Sanh, Chaumond, Delangue, Moi, Cistac,
  Rault, Louf, Funtowicz, Davison, Shleifer, von Platen, Ma, Jernite, Plu, Xu,
  Le~Scao, Gugger, Drame, Lhoest, and Rush}]{wolf2019huggingface}
Thomas Wolf, Lysandre Debut, Victor Sanh, Julien Chaumond, Clement Delangue,
  Anthony Moi, Pierric Cistac, Tim Rault, Remi Louf, Morgan Funtowicz, Joe
  Davison, Sam Shleifer, Patrick von Platen, Clara Ma, Yacine Jernite, Julien
  Plu, Canwen Xu, Teven Le~Scao, Sylvain Gugger, Mariama Drame, Quentin Lhoest,
  and Alexander Rush. 2020.
\newblock \href {https://doi.org/10.18653/v1/2020.emnlp-demos.6} {Transformers:
  State-of-the-art natural language processing}.
\newblock In \emph{Proceedings of the 2020 Conference on Empirical Methods in
  Natural Language Processing: System Demonstrations}, pages 38--45, Online.
  Association for Computational Linguistics.

\bibitem[{Zhang et~al.(2020)Zhang, Liu, Pan, Song, and Leung}]{ZhangLPSL20}
Hongming Zhang, Xin Liu, Haojie Pan, Yangqiu Song, and Cane~Wing{-}Ki Leung.
  2020.
\newblock \href {https://doi.org/10.1145/3366423.3380107} {{ASER:} {A}
  large-scale eventuality knowledge graph}.
\newblock In \emph{{WWW} '20: The Web Conference 2020, Taipei, Taiwan, April
  20-24, 2020}, pages 201--211. {ACM} / {IW3C2}.

\bibitem[{Zhao et~al.(2020)Zhao, Walker, and Chaturvedi}]{zhao2020bridging}
Chao Zhao, Marilyn Walker, and Snigdha Chaturvedi. 2020.
\newblock \href {https://doi.org/10.18653/v1/2020.acl-main.224} {Bridging the
  structural gap between encoding and decoding for data-to-text generation}.
\newblock In \emph{Proceedings of the 58th Annual Meeting of the Association
  for Computational Linguistics}, pages 2481--2491, Online. Association for
  Computational Linguistics.

\bibitem[{Zhou et~al.(2021)Zhou, Lee, Selvam, Lee, and
  Ren}]{zhou2021pretraining}
Wangchunshu Zhou, Dong-Ho Lee, Ravi~Kiran Selvam, Seyeon Lee, and Xiang Ren.
  2021.
\newblock \href {https://openreview.net/forum?id=3k20LAiHYL2} {Pre-training
  text-to-text transformers for concept-centric common sense}.
\newblock In \emph{International Conference on Learning Representations}.

\end{thebibliography}
\bibliographystyle{acl_natbib}

\appendix

\section{Training Details}
\label{app::train}
The BART and T5 models are implemented using the Transformers library \cite{wolf2019huggingface}. We fine-tune each model on the training data of \textsc{CommonGen} with Adam \cite{kingma2014adam}. We set the learning rate as 2e-5 and adopt early stopping based on the loss of development set. The batch size of training is 64.

\section{Human Evaluation Details}
\label{app::eval}
We randomly select 100 test instances that had 5 concepts as input, since they are more challenging than those with fewer concepts. The three measures we used are 1)  Rationality: whether or not the sentence is in accordance with commonsense; 2) Fluency: whether or not the sentence is fluent and has no grammatical errors; and 3) Succinctness: whether or not the sentence contains redundant words or repeated information. 

The pairwise scores of those measures are calculated as follows. When comparing a certain approach to Planned-BART \textsubscript{Rank}, we report the percentage of instances that were judged to be better/worse/same than those of Planned-BART \textsubscript{Rank}, yielding a score ranging from -1 (unanimously worse) to 1 (unanimously better). For example, when evaluating the rationality scores, Unordered-BART \textsubscript{Vanilla} performs better/worse/same than Planned-BART \textsubscript{Rank} for 27\%/65\%/8\% of the instances, yielding a pairwise score as 0.27-0.65=-0.38.

\section{Permutation Invariance}
\label{app::invariance}
To figure out to what extent Unordered-BART is permutation-invariant, we conduct the following analysis. For each test instance, we feed all different input permutations to the Unordered-BART to obtain the corresponding output sentences. We measure the invariance of the outputs by computing the percentage of the most frequent output. If the concept order does not affect the output, all different permutations will lead to identical outputs and the percentage will be 100\%. If half of the permutations obtain the same outputs, the percentage will be 50\%. We measure the input invariance at the following two levels.

\noindent\textbf{Sentence-level invariance} For various permutations of a specific concept-set, if the percentage of the most frequent sentence is greater than an invariance threshold $\alpha$, we regard the model as input-invariant to that instance at the sentence level. 

\noindent\textbf{Skeleton-level invariance}
The sentence-level invariance requires the output sentences to be identical, which is strict and does not consider minor lexical differences (e.g., the difference in function words or modifiers). Therefore, we also report the skeleton-level invariance by measuring the percentage of the most frequent skeleton (mode skeleton). It reflects whether or not the model output will follow a certain order under different permutations, which is a more forgivable invariance measure compared to its sentence-level counterpart.

\begin{figure}[H]
     \centering
     \begin{subfigure}[b]{0.23\textwidth}
         \centering
         \includegraphics[width=\textwidth]{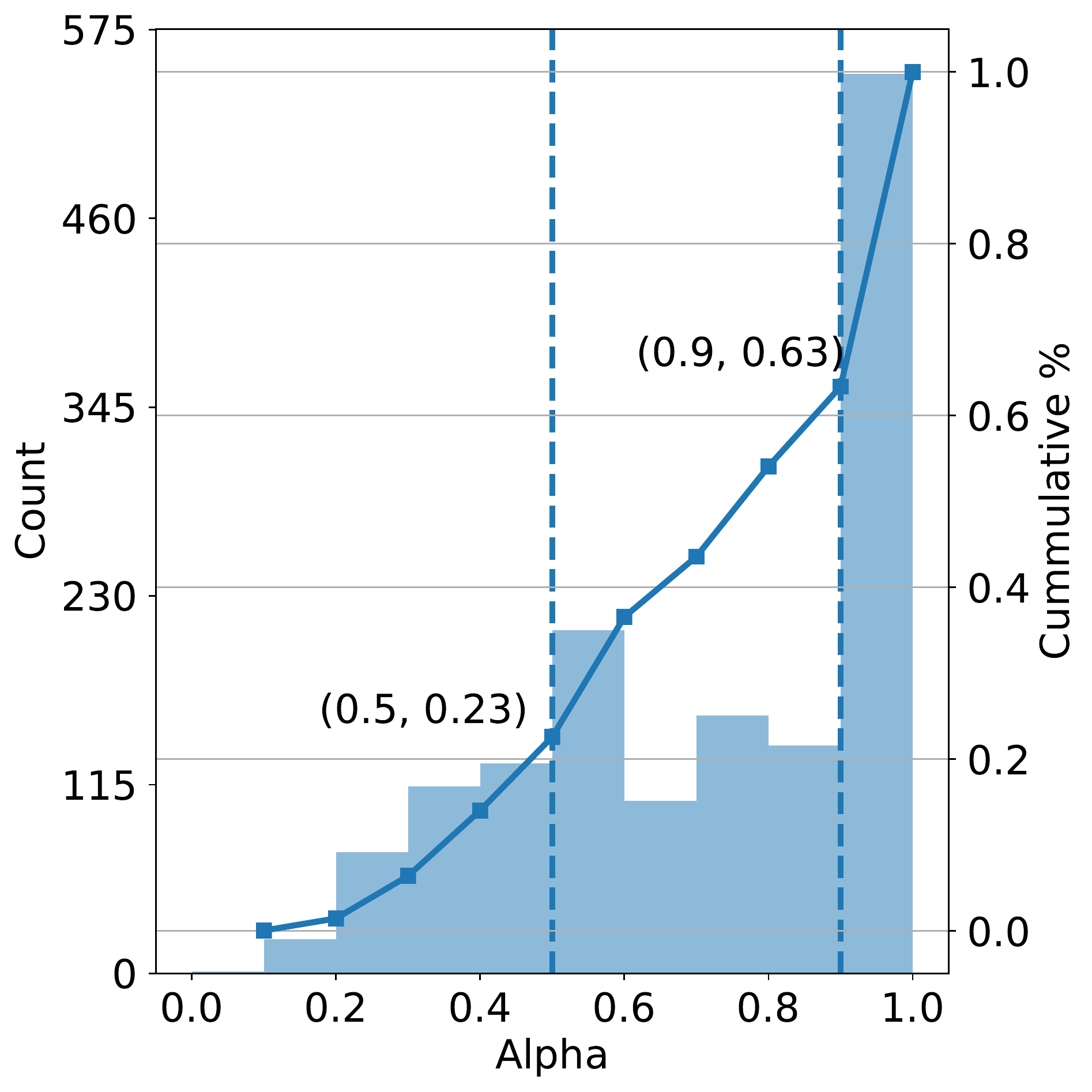}
     \end{subfigure}
     \hfill
     \begin{subfigure}[b]{0.23\textwidth}
         \centering
         \includegraphics[width=\textwidth]{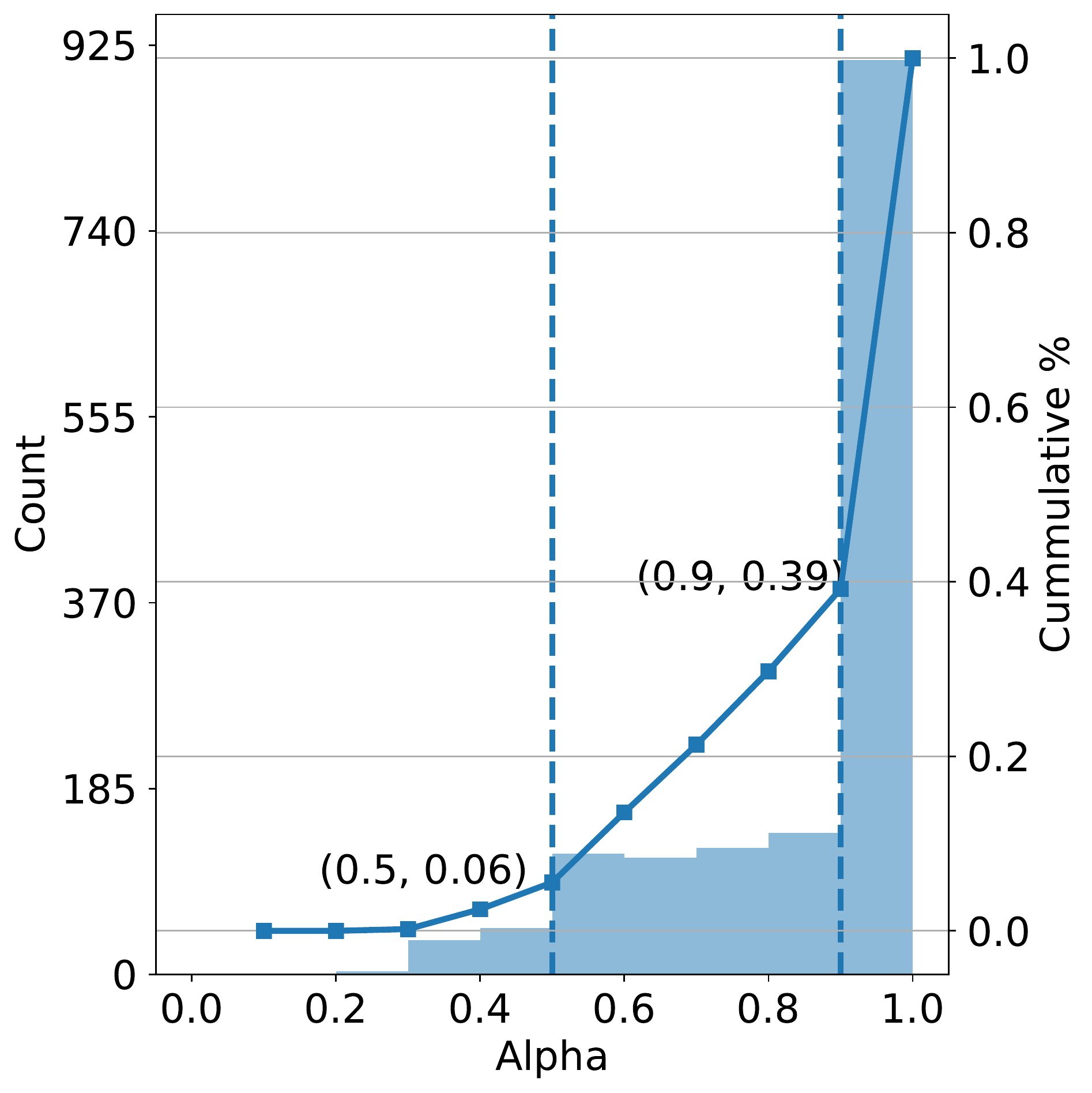}
     \end{subfigure}
\caption{The histogram of \textsc{CommonGen} test set w.r.t. the percentage of most frequent sentences (left) and skeletons (right), respectively. We also show the cumulative distribution in blue lines. When $\alpha=0.9$, 37\% and 61\% of the test instances are invariant at the sentence and skeleton-level, respectively.}
\label{fig::invariance}
\end{figure}

Figure \ref{fig::invariance} shows the distribution of \textsc{CommonGen} test set w.r.t. the percentage of most frequent sentences (left) and skeletons (right). 
When setting $\alpha=0.9$
, 37\% of the test instances are invariant at the sentence level, and 61\% are invariant at the skeleton level. This indicates that for 61\% of the permutations, Unordered-BART can organize the concepts in one particular order irrespective of the input order.

\section{Analysis of Local Attention}
\label{app::local_attn}
In addition to the analysis of global attention distribution, we also investigate local attention patterns, i.e., the attention weights between concept tokens. Previous works show that some attention heads can reflect certain aspects of syntactic and semantic relations between words \cite{clark2019does, htut2019attention}. We want to investigate if the heads can track commonsense relations between concepts.

For this purpose, we first build gold relations between the concepts that capture commonsense knowledge. One option is to utilize ConceptNet relations between concepts~\cite{lin2020commongen}. However, these relations connect only two concepts at a time disregarding the context information from other concepts. Consider \{throw, catch, dog, frisbee\} as an example. 
``Dog'' may be ``caught'' but this relation is less plausible in this case because of the existence of ``frisbee''. When considering this context, humans provide references such as ``The dog catches the frisbee when the boy throws it.''

Another option is to use the dependency relations between words in the reference sentences, which can capture the commonly occurring relations between concepts while incorporating the context. For example, the relation ``catch$\xrightarrow{\text{dobj}}$frisbee'' captures the commonsense that frisbee is often caught,
Similar ideas are also adopted in \citet{ZhangLPSL20}. In particular, we extract the one-hop and two-hop dependency relations of all concept pairs from the references, and only keep the relations that appear in two or more references. 

\begin{table}
		\centering
		\small
		\begin{tabular}{l|c|c|c } 
 \hline
  \textbf{Relation} &\textbf{Head} &\textbf{UAS} & $\bm{I}_{lh}$ \\ 
 \hline
v-dobj-n &  10-7 & 83.97 & 13\\
v-prep-adp-pobj-n & 10-7 & 82.61 & 13\\
n-nsubj-v & 11-12 & 87.62 & 3\\
n-prep-adp-pobj-n &  10-8 & 60.55 & 15\\
v-advcl-v & 4-11 & 85.37 & 4\\
v-conj-v & 10-1 & 83.07  & 2\\
n-nsubj-v-dobj-n & 6-16 & 62.34 & 40 \\
v-xcomp-v & 8-15 & 91.32 & 19\\
n-conj-n &   10-0 & 88.75 & 56\\
n-comp-n &   1-8 & 83.31 & 24\\
 \hline
\end{tabular}

		\caption{\label{tab::head_data} The functional heads for each relation, as well as the corresponding UAS and importance rank.}
\end{table}

For attention probing, we use attention weights between input tokens to reflect the strength of their associations. Given two concepts $c_i$ and $c_{j, j>i}$, we regard them as strongly associated under the attention head $(l, h)$ if the attention weight $\alpha^{l,k}_{ij}$ is the highest among the scores from all the other concepts $c_{k\backslash \{i,j\}}$ to $c_j$.

\begin{figure}[t]
    \centering
    \includegraphics[width=1\linewidth]{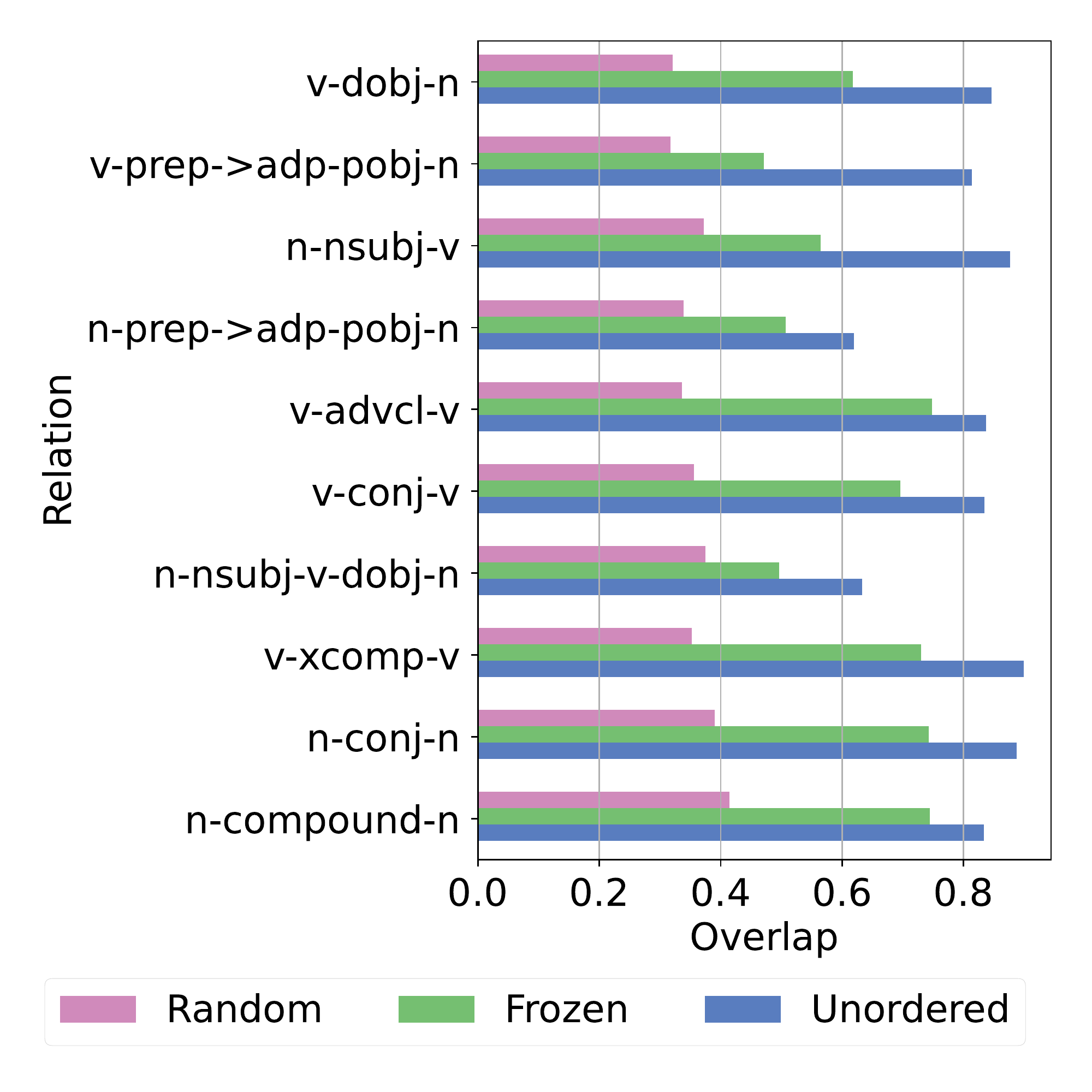}
    \caption{UAS of commonsense relations from three BART models via attention probing. The performance of fine-tuned Unordered-BART > pre-trained Frozen-BART > randomly-initialized BART among all of the relations. }
    \label{fig:rel-attn-acc}
\end{figure}

\begin{table*}[t!]
		\centering
		\small
		\begin{tabularx}{\linewidth}{@{\hspace{0pt}}p{3cm}p{8cm}X@{\hspace{0pt}}}
	\hline 
	Model & Output & Skeleton \\\hline
	Unordered-BART & A crowd of people watch and dance to the music. & crowd watch dance music \\\hline
	Planned-BART &   A crowd of people are dancing to music while others watch. & crowd dance music watch \\
	& A man plays music and watches the crowd dance. & music watch crowd dance \\
	 &  A group of people dance to music as a crowd watches. & dance music crowd watch \\
& 	A man watches a crowd of people dancing to music. & watch crowd dance music \\\hline
	Human & The crowd likes to watch her dance to the music. & crowd watch dance music  \\
& 	The crowd watched the dance, and listed to the music. & crowd watch dance music \\
	& I watched as the crowd dance to the music. & watch crowd dance music \\
& 	 A person dancing to the music as a crowd of people watch. & dance music crowd watch \\ \hline
\end{tabularx}

		\caption{\label{tab::example} Sample texts generated by Unordered-BART, Planned-BART, and humans for the concept set \{dance, music, crowd, watch\}. The diversity of Planned-BART is more close to human generation.}
\end{table*}

\begin{figure}
     \centering
     \includegraphics[width=1\linewidth]{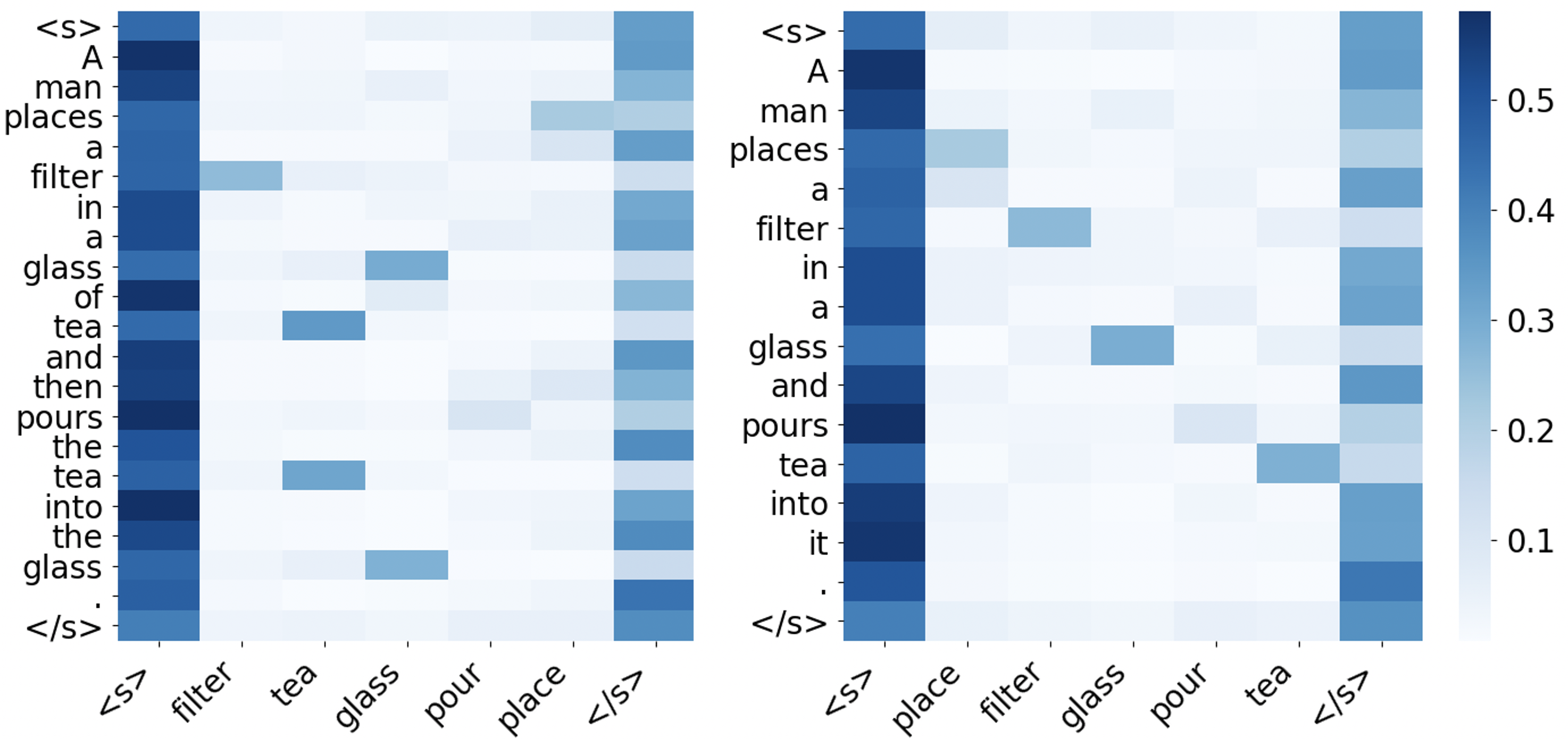}
\caption{The cross-attention matrix of two permutations of the same concept-set produced by Unordered-BART. It's difficult for Unordered-BART to learn the optimal order of attention.}
\label{fig::cross-att}
\end{figure}

We choose the 10 most common dependency relations from the test set and report the Unlabeled Attachment Score (UAS) %
of attention probing in Figure \ref{fig:rel-attn-acc}. We also list the UAS of a randomly initialized BART and a pre-trained BART without fine-tuning for comparison.

Results in Figure \ref{fig:rel-attn-acc} show that for some frequent and simple relations such as ``v-dobj-n'' and ``n-nsubj-v'', there is at least one attention head that tends to track them regardless of the differences in the permutation orders. For example, attention head ``Layer-10 Head-7'' tracks the ``v-dobj-n'' relation with a UAS of 83.9\%. The comparison among the three models shows that the pre-trained BART already exceeds the randomly initialized model in tracking commonsense relations between words, and fine-tuning further strengthens those relations. 

To demonstrate that these functional heads are important for generation, we use the expected sensitivity \cite{NEURIPS2019_2c601ad9} of the model to each head to evaluate the head importance as 
\begin{equation}
I_{lh}=\mathbb{E}_{x \sim X}\left|\frac{\partial \mathcal{L}(x)}{\partial \xi_{lh}}\right|
\end{equation}
where $\mathcal{L}(x)$ is the loss of generation and $\xi_{lh}$ is the mask variable for head $l-h$ with values in $\{0, 1\}$. The general idea is that the value change of important heads can have a larger impact on the model loss. Results are shown in Table \ref{tab::head_data}. For most relations, the corresponding functional heads also have a high rank of importance. This consistency indicates that the model heavily relies on these heads when generating texts, and further demonstrates that the finetuned BART can capture the commonsense between concepts for generation.

\section{Impact on Repetition}
\label{app::rep}
The repetition of the unordered BART is caused by the order-agnostic property of its input. Since the input concepts are unordered, the decoder cannot pay attention to the input in a monotonic way (from left to right) during decoding, which may mislead the decoder to attend to the concepts that have been previously generated. For example, on the left of Figure \ref{fig::cross-att}, the decoder attends to ``tea'' and ``glass'' twice during decoding, which achieves the local coherence but causes the global repetition issue and unnatural text. However, when modifying the input in another order, as shown in the right of Figure \ref{fig::cross-att}, the repetitive and unnatural expressions disappear. It indicates that the BART decoder has difficulty ordering the input globally, and providing a well-ordered plan as input can alleviate this issue.
On the contrary, in Planned-BART, the decoder can assign attention weights monotonically to the input, and therefore reduce the repetition caused by re-attending the previous concepts.

\section{Impact on Diversity}
\label{app::div}
Table \ref{tab::example} provides an example with the outputs created by both models and humans. Unordered-BART can create only one output due to the permutation invariance. Also, the object of \textit{watch} is missing in its output. On the other hand, similar to the human-written output, the output of Planned-BART is more natural and diverse.

\end{document}